\documentclass{article}
\usepackage{spconf,amsmath,graphicx}
\usepackage{multirow}

\title{SEMI-SUPERVISED LEARNING FOR ROBUST SPEECH EVALUATION}
\name{Huayun Zhang, Jeremy H.M. Wong, Geyu Lin, Nancy F. Chen}
\address{Institute for Infocomm Research (I2R), Agency for Science, Technology and Research (A*STAR)\\ 1 Fusionopolis Way, Connexis \#21-01, Singapore 138632}
\begin{document}
%
\maketitle
\begin{abstract}
Speech evaluation measures a learner’s oral proficiency using automatic models. Corpora for training such models often pose sparsity challenges given that there often is limited scored data from teachers, in addition to the score distribution across proficiency levels being often imbalanced among student cohorts. Automatic scoring is thus not robust when faced with under-represented samples or out-of-distribution samples, which inevitably exist in real-world deployment scenarios. This paper proposes to address such challenges by exploiting semi-supervised pre-training and objective regularization to approximate subjective evaluation criteria. In particular, normalized mutual information is used to quantify the speech characteristics from the learner and the reference. An anchor model is trained using pseudo labels to predict the correctness of pronunciation. An interpolated loss function is proposed to minimize not only the prediction error with respect to ground-truth scores but also the divergence between two probability distributions estimated by the speech evaluation model and the anchor model. Compared to other state-of-the-art methods on a public data-set, this approach not only achieves high performance while evaluating the entire test-set as a whole, but also brings the most evenly distributed prediction error across distinct proficiency levels. Furthermore, empirical results show the model accuracy on out-of-distribution data also compares favorably with competitive baselines.

\end{abstract}
\section{Introduction}
\label{sec:intro}

Speech evaluation is traditionally performed by human experts. Recently, machine learning models that correlate well with human judgement have made automated evaluation possible - predicting consistent scores with less human effort \cite{Dalen2015AutomaticallyGL}.

Early speech assessment studies have focused on crafting discriminative features and testing with different statistical models. These features include duration-related features \cite{596227}, goodness of pronunciation (GOP) and its variants \cite{WITT200095, Huang2017ATL}, linguistic features \cite{Shi2020ContextawareGO}, the marginal distribution of speech signals \cite{Cheng2020ASRFreePA} and much more. Early modelling works focused on statistical models such as Gaussian Processes \cite{Dalen2015AutomaticallyGL}, while more recent works have focused on deep learning approaches. \cite{HU2015154}  extended GOP to Deep Neural Networks (DNN) trained acoustic model and improved mispronunciation detection performance. 
\cite{Gretter2019AutomaticAO} used a similar approach by also taking advantage of language models - a sequence-trained acoustic model and multiple n-gram language models were used to compute acoustic and text-based features for a final neural classifier. \cite{wang20ba_interspeech} and \cite{Wang2021AutomatedSO} examined the use of Transformer-style scoring models based solely on transcriptions. They reported strong performance approaching or surpassing human agreement in speech evaluation.

A fundamental challenge to deep learning algorithms is the
reliance on supervised learning and by extension good data, meaning data-sets that are large, representative and balanced. Small, imbalanced data-sets often lead to over-fitting and issues with generalizability when models face out-of-distribution (OOD) or under-represented samples \cite{Heimbalance}. While not unique to speech
evaluation, the characteristics of speech evaluation data often possess long-tail distributions stemming from mispronunciations and accents from non-native speakers, in addition to inter-rater variances from teacher scores.

Different learning strategies have been studied to address
this challenge in speech evaluation. \cite{Malinin2019UncertaintyEI} and \cite{ wu20n_interspeech} investigated quantifying the assessment uncertainty using ensemble, distillation and multi-task learning. Improved scoring performance and robustness to OOD was reported. \cite{Yang2014MachineLA} 
investigated methods for cost-sensitive learning and synthetic minority over-sampling and reported significantly better pronunciation error detection rates on imbalanced corpora. \cite{wong22} explored supervised multi-task ensemble learning as a way to exploit mutual information between different annotation types.

Recently, a self-attention based Transformer encoder model - GOPT \cite{gopt} was applied to speech evaluation. Leveraging self-supervised learning, GOPT and large-scale pre-trained universal speech embedding has demonstrated superior performance on a speech evaluation data-set \cite{gopt,3m} . However, these studies only reported  performance improvement when the entire testing data was taken as a whole. While investigating the cohort-wise performance, we found that its prediction quality deteriorated significantly for minority cohorts. In addition, the model’s robustness has not been verified in cross-data-set test.

Robustness issues will cause consistency and fairness concerns for automatic speech evaluation. This study focuses on improving model's robustness while training data is imbalanced and insufficient. A two-stage training is proposed to prevent over-fitting on small and biased training samples and to enhance model's generalizability in real world. An objective metric is adopted to calculate pseudo-scores for speech evaluation. An anchor model is pre-trained in semi-supervised manner using these pseudo-scores. While training the real speech evaluation model, the prediction made by this anchor model is used to interpolate the loss function. We will report both overall and cohort-wise performance, as well as the result for a cross-data-set test.

\section{Architecture}
\label{sec:format}

\subsection{Speech Evaluation Model}
\label{ssec:subhead}

In reading speech evaluation, evaluators assess the individual's language proficiency according to a rubric containing the criteria and standards to be used,  assessing different aspects of proficiency. 
In automatic evaluation, the evaluators are replaced by a model trained on human labelled samples.

\begin{equation}
    {y}=\Phi(\mathbf{O},T)
\end{equation}
 The input is a pair of speech and reference. $\mathbf{O}=[\mathbf{o}_1,...,\mathbf{o}_m]$ is the sequence of $m$ consecutive speech observations. $T=[w_1,...,w_M]$ is the sequence of $M$ consecutive canonical-phonemes in the reading text, either known a priori or generated by Automatic Speech Recognition (ASR).  $\Phi$ is the model. $y$ is the numerical score associated with the input tuple $(\mathbf{O},T)$. To compare a speaker's pronunciation to a certain level of expected proficiency and identify errors in the sentence, $\mathbf{O}$ has to be aligned to $T$. Model training is simplified by extracting phone-wise features $\mathcal{F}([\mathbf{o}]_{w_{i}},w_i)$ along the sentence, where $[\mathbf{o}]_{w_{i}}$ is the speech segment belonging to $w_i$.

 \begin{equation}
{y}=\Phi\left(\left[\begin{array}{cc}
\mathcal{F}([\mathbf{o}]_{w_{1}},w_1)\\
\vdots\\
\mathcal{F}([\mathbf{o}]_{w_{i}},w_i)\\
\vdots\\
\mathcal{F}([\mathbf{o}]_{w_{M}},w_M)
\end{array}
\right]\right)\\
\end{equation}

\subsection{Speech Evaluation Feature}
An assembly of traditional speech evaluation features and cutting-edge embedding features will be studied in this paper.

Traditional speech evaluation features are a set of indicators specially designed to capture the acoustic, temporal, and linguistic cues in the reading speech, including:

\begin{itemize}
\item{\textit{GOP}}\\
It was proposed in \cite{WITT200095} and its version with deep neural networks was introduced in \cite{HU2015154}. 
\item{\textit{Tempo}} \\
This feature measures the dynamic temporal variation in the pronunciation. It was introduced in \cite{Zhang2021MultilingualSE}.
\item{\textit{PhoEmb}} \\
Phonetic Embedding provides semantic information for speech evaluation. It was introduced in \cite{Zhang2021MultilingualSE}.
\item{\textit{Pitch}}\\
The average values of frame-wise pitch and delta-pitch for each canonical-phoneme intervals \cite{Zhang2021MultilingualSE}.
\end{itemize}

A set of state-of-the-art Transformer models were downloaded from Huggingface \cite{huggingface}. A forward pass is done using these pre-trained models and their encoder outputs are adopted as frame-wise speech representation. Average aggregation is conducted at canonical-phoneme intervals to get phone-wise evaluation embedding  $\mathcal{F}^{pre}$. These pre-trained Transformer models include: 
\begin{itemize}
\item{\textit{"facebook/wav2vec2-large-xlsr-53"}} \cite{wav2vec2}\\
It was trained on more than 50k hours of unlabelled speech in 53 languages, including Multilingual Librispeech \cite{mls}, CommonVoice \cite{commonvoice}, and Babel \cite{babel}.
\item{\textit{"facebook/hubert-large-ll60k"}} \cite{hubert}\\
It was trained on Libri-Light 60k hours data \cite{librilight}.
\item{\textit{"microsoft/wavlm-large"}} \cite{wavlm}\\
It was trained on mix 94k hours of data, including 60k hours Libri-Light, 10k hours GigaSpeech \cite{gigaspeech}, and 24k hours VoxPopuli \cite{voxpopuli}.
\item{\textit{"openai/whisper-large-v2"}} \cite{whisper} (encoder only)\\ 
It was trained on 680k hours of multilingual and multitask supervised data collected from the web.
\end{itemize}

All features are concatenated together at corresponding canonical-phonemes. 
\begin{equation}
\begin{cases}
\mathcal{F}(\cdot)&=[GOP,Tempo,PhoEmb,Pitch,\mathcal{F}^{pre}]\\
\mathcal{F}^{pre}&=[\mathcal{H}_{wav2vec2},\mathcal{H}_{hubert},\mathcal{H}_{wavlm},\mathcal{H}_{whisper}]\\
\end{cases}
\end{equation}

\begin{figure*}[ht!]
  \centering
  \includegraphics[width=\linewidth, scale=0.08]{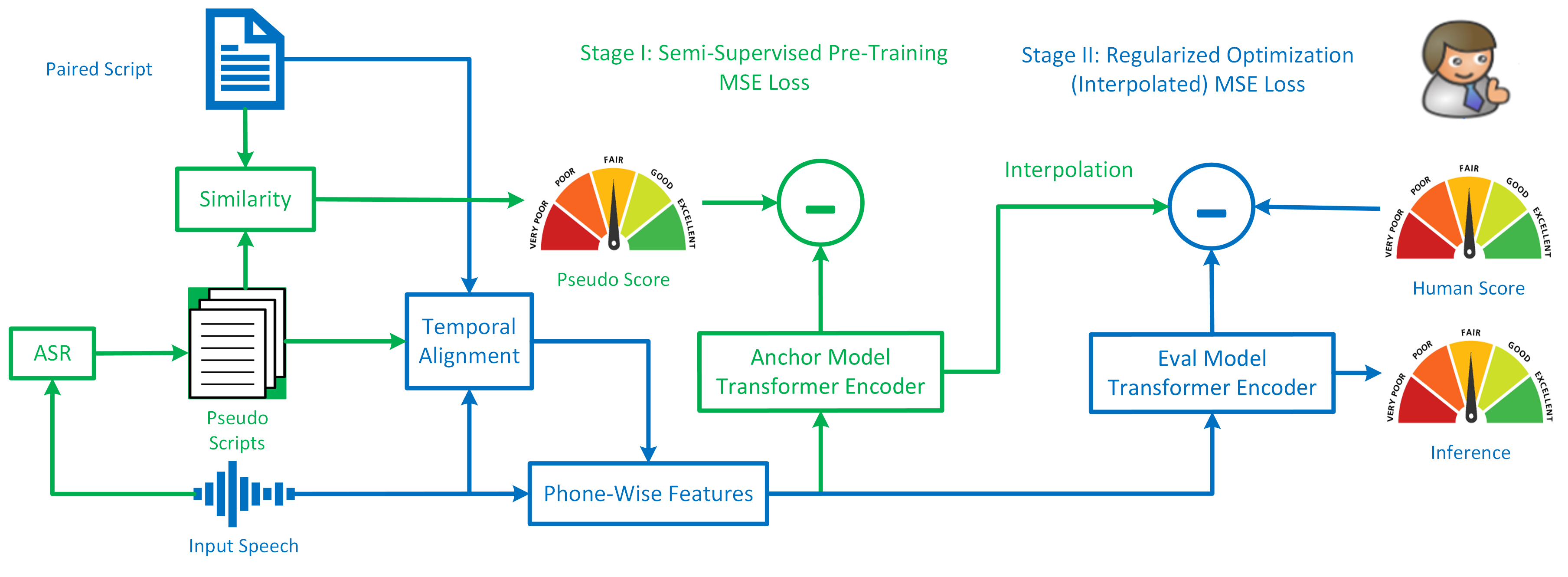}
  \vspace{-1.1 cm}
  \caption{Proposed System Diagram.}
  \label{fig:Proposed 2 step}
\end{figure*}
\section{Two Stage Model Training}
Figure 1 is the diagram of the proposed two-stage training. In the first stage, an anchor model is trained on augmented data with pseudo-scores reflecting the similarity between speech and reference. In the second stage, the prediction made by the anchor model is used to interpolate the training loss and prevent evaluation model predicting scores too different than the underlying similarity between input speech and reference. In Fig.\ref{fig:Proposed 2 step}, the green parts are the extra function modules introduced for semi-supervised pre-training (the stage-I training) and regularized optimization (the stage II training). At inference time, the system works using only the blue modules. The evaluation model has the same GOPT structure as the models studied in \cite{gopt} and \cite{3m}. 
\subsection{Objective Metric}
\label{ssec:obj}

The difficulty in collecting accurate human-scores for speech assessment motivates our searching for a metric that can capture speech correctness without subjective bias. An automatic computation of such a metric can be used to assign pseudo-scores for a large amount of speech data without accurate human-scores, thereby increasing the training data coverage.

From Shannon's theory of communication \cite{shannon}, speech evaluation can be modelled as a noisy communication channel. The reference $T$ is encoded into voice by the speaker (encoder) and sent to the evaluator (decoder). The voice is received and transcribed (decoded) into text $\hat{T}$. Mutual information is often used to measure the information transmitted through the communication channel. It reflects the similarity between the input $T$ and output $\hat{T}$. It is estimated by sampling the phone distributions of these two random variables.

\begin{equation}
    I(T,\hat{T}) = \sum_{i=1}^{C}\sum_{j=1}^{C}{p(w_i,w_j)\log\frac{p(w_i,w_j)}{p(w_i)p(w_j)}}
\end{equation}
where $w_i$ is the $i^{th}$  phone to be sampled and $C$ is cardinality of the phone-set determined by data $T \bigcup{\hat{T}}$. ${p(w_i, w_j)}$ is the joint distribution between $T$ and $\hat{T}$. 
 $p(w_i)$ and $p(w_j)$ are two marginal distributions. These distributions could be approximated using a confusion matrix. In this study, a sentence-level phonetic confusion matrix is constructed by a dynamic programming between each pair of $(T,\hat{T})$. (A place holder $*$ is added into the phone-set to indicate the insertions and deletions  in the alignment between $T$ and $\hat{T}$, when necessary.)

 $I(T,\hat{T})=0$ if $T$ and $\hat{T}$ are independent. $I(T,\hat{T})$ is non-negative and a higher mutual information indicates a larger reduction in uncertainty about one given another. When $T$ and $\hat{T}$ are identical, the mutual information will be the maximum possible value $\log(C)$. $C$ is data-dependent.  Data-dependent upper bound  makes it unsuitable as an evaluation metric. Instead, normalized mutual information \cite{e19110631} is adopted.
 \begin{equation}
\label{ni}
\begin{split}
NI(T,\hat{T}) &= \frac{2\times I(T,\hat{T})}{H(T)+H(\hat{T})}\\
H(T) &= \sum_{i=1}^{m}{p(w_i)\log(1/p(w_i)}\\
H(\hat{T}) &= \sum_{j=1}^{n}{p(\omega_j)\log(1/p(\omega_j))}
\end{split}
\end{equation}
$H(T)$ and $H(\hat{T})$ are the entropy (uncertainty) of variables. 
$0\leq {NI(T,\hat{T})} \leq{1}$. 
When $I(T,\hat{T})=H(\hat{T})=H(T)$, $NI(T,\hat{T})=1$. It means that all knowledge about $\hat{T}$ comes from $T$, no information loss and no extra noise introduced.

\subsection{Stage-I: Semi-Supervised Pre-training}

To enhance data diversity,  training sample $(\mathbf{O},T)$ without human-score can be automatically labelled using Eq.\eqref{ni}, where $\hat{T}$  is generated by ASR in stead of human evaluator. A new 3-tuple training sample $(\mathbf{O},T,\hat{T})$ is constructed by this method. As $\mathbf{O}$ is perceived as  $\hat{T}$ (instead of $T$), the phone-wise evaluation feature is extracted according to $\hat{T}$.  Eq.\eqref{ni} defines a measurement on the similarity between speech $\mathbf{O}$ (ground-truth reference is $T$) and its pseudo reference \footnote{we use "pseudo reference" to denote that phone-wise features are calculated with respect to the ASR n-best instead of the ground truth. Normally, phone-wise features are calculated with respect to ground-truth reference.} $\hat{T}$. A pseudo-score $\hat{s}$ is assigned according to Eq\eqref{ni} for $(\mathbf{O},T,\hat{T})$  to measure  the correctness of $\mathbf{O}$  with associated $\hat{T}$. Multiple pseudo references $\hat{T}$  are taken from ASR n-best transcriptions and multiple $(\mathbf{O},T,\hat{T})$ are spawned from $(\mathbf{O},T)$. Pre-training  leverages the enriched variation in 3-tuples to uncover complementary  patterns between speech and reference.

While data augmentation increases data diversity and therefore can improve model robustness \cite{robust}, it's difficult to scale it to a specific training task. Inappropriate data combination leads to under-fitting and poor performance. To avoid augmented data dominating the model training, data augmentation is not directly applied while training the real evaluation model. Instead, a separate anchor model is trained using augmented samples $(\mathbf{O},T,\hat{T})$ in a semi-supervised manner. Prediction made by this anchor model is then added as a regularization term to the loss function in stage-II  training.

\subsection{Stage-II: Regularized Optimization} %
While training the speech evaluation model, the cross-entropy loss is  aggregated over the training samples.
\begin{equation}
    D=-\frac{1}{N}\sum_{t=1}^N\sum_{s=1}^{S}{\Tilde{p}(y|\mathbf{O}_t,T_t)\log[p(y|\mathbf{O}_t,T_t)]}
    \label{6}
\end{equation}
where $y$ is the predicted score given the speech $\mathbf{O}_t$ and associated reference $T_t$. $\Tilde{p}(y|\mathbf{O}_t,T_t)$ is the target probability for $(\mathbf{O}_t,T_t)$. $p(y|\mathbf{O}_t,T_t)$ is the model estimated probability for $(\mathbf{O}_t,T_t)$. $N$ is the number of samples in the batch and $S$ is the total number of marking bands. In most cases, samples are human-labelled, i.e., given the human-score $s_t$ for sample $(\mathbf{O}_t,T_t)$, $\Tilde{p}(y|\mathbf{O}_t,T_t)=1$ when $y=s_t$ and $\Tilde{p}(y|\mathbf{O}_t,T_t)=0$ otherwise. Cross-entropy loss could be simplified as: 
\begin{equation}
    D=-\frac{1}{N}\sum_{t=1}^N\Tilde{p}(y=s_t|\mathbf{O}_t,T_t)\log[p(y,s_t|\mathbf{O}_t,T_t)]
\label{7}
\end{equation}
Minimizing the cross-entropy loss on a small and biased data will make the model over-fit and lead to poor generalisation capability. To prevent this, a regularization term between the probabilities estimated from the evaluation model and a separate anchor model is added into the loss function:
\begin{equation}
\overline{D}=D+\rho\cdot D_{KL}(p(y|\mathbf{O}_t,T_t)\Vert q(y|\mathbf{O}_t,T_t))\\
\label{8}
\end{equation}
where $q(y|\mathbf{O}_t,T_t)$ is the probability estimated from anchor model for $(\mathbf{O}_t,T_t)$. $D_{KL}(p\Vert q)=\frac{1}{N}\sum_{t=1}^Np\cdot\log(q/p)$ is the Kullback-Leibler-Divergence (KLD) between two predictive distributions. $0\leq\rho\leq 1$ is a hyper-parameter. $D_{KL}(p\Vert q)$ is the expectation of logarithmic difference between $p$ and $q$, where the expectation is taken using $p$. Expand the KLD term in Eq.\eqref{8} and substitute the term $D$  with Eq.\eqref{7}. 
\begin{equation}
\begin{split}
    \overline{D}=-\frac{1}{N}\cdot&\sum_{t=1}^N\Tilde{p}(y=s_t|\mathbf{O_t},T_t)\log[p(y,s_t|\mathbf{O}_t,T_t)]\\
    -\frac{\rho}{N}\cdot&\sum_{t=1}^N{p(y=\hat{s}_t|\mathbf{O}_t,T_t)\log[q(y,\hat{s}_t|\mathbf{O}_t,T_t)]}\\
    +\frac{\rho}{N}\cdot&\sum_{t=1}^N{p(y=s_t|\mathbf{O}_t,T_t)\log[p(y,s_t|\mathbf{O}_t,T_t)]}
\end{split}
\label{9}
\end{equation}
where $\hat{s}_{t}$ is the pseudo-score assigned to $(\mathbf{O}_t,T_t)$ and it is computed by a forward pass using  the anchor model. As a grading machine, models' predictions are expected to be normally-distributed.  Assuming ${p(y,s_t|\mathbf{O}_t,T_t) \sim \mathcal{N}(s_t,\sigma)}$ and ${q(y,\hat{s}_t|\mathbf{O}_t,T_t) \sim \mathcal{N}(\hat{s}_t,\sigma)}$, the KLD regularized loss $\overline{D}$ becomes: ( constants unrelated to the model are removed.) 
\begin{equation}
\begin{split}
\overline{D} = -\frac{1-\rho}{N}\cdot&\sum_{t=1}^N(y-s_t)^2\\
-\frac{\rho}{N}\cdot&\sum_{t=1}^N[\exp(-(\hat{s}_t-s_t)^2/2)\cdot(y-\hat{s}_t)^2]
\end{split}
\label{imse}
\end{equation}
where $\exp(-(\hat{s}_t-s_t)^2/2)$ is the Gaussion kernal distance between $s_t$ and $\hat{s}_t$.

If $\rho=0$ or $\hat{s}_t=s_t$ for all $t\in [1,...,N] $, 
minimizing   $\overline{D}$ of Eq.\eqref{8} is equivalent to the minimization of  MSE loss with respect to human-scores (the first term on the right side of Eq.\eqref{imse}). If $\rho=1$, Eq.\eqref{imse} reduces to weighted MSE (WMSE) loss with respect to pseudo-scores (the second term on the right side of Eq.\eqref{imse}). If $0<\rho<1$, 
minimizing the KLD regularized cross-entropy loss of Eq.\eqref{8} is equivalent to the minimization of a distance weighted interpolation between MSE loss with respect to human-scores and WMSE loss with respect to pseudo-scores. Interpolated MSE loss (iMSE)  prevents model from predicting scores very different than the underlying similarity between speech and reference. 
\section{Experiments}
\subsection{ASR}
A TDNN \cite{tdnn} acoustic model with 20M parameters was trained on Librispeech \cite{librispeech} using Lattice-Free MMI \cite{Povey2016PurelySN}. It is used to calculate the time stamps for canonical-phonemes.

This model is also used together with a librispeech 3-gram language model \cite{librispeech} to get the n-best transcriptions in data augmentation and stage-I anchor model training.

\subsection {Data}
\label{ssec:data}
\begin{itemize}

\item{\textbf{Evaluation Data with Human Scores}}\\
\textit{SpeechOcean762} \cite{762} is a publicly available  data-set for English pronunciation assessment. It has 5,000 English sentences collected from 250 Mandarin speakers, including children and adults. To avoid subjective bias and label the data in a consistent manner, each sentence is evaluated by 5 language experts independently and is allotted a score between 1-10 (1 being the lowest, and 10 being the highest) for 3 aspects of the language proficiency, pronunciation(accuracy), rhythm(fluency), and intonation(prosodic). The median score from these 5 human experts is used as the final score of the sentence. This data comes with a training and testing split, 2,500 sentences (125 speakers) each. There is no script overlap or speaker overlap between training and testing.

\item{\textbf{Augmented Data for Anchor Model Training}}\\
Augmented training samples $(\mathbf{O},T,\hat{T})$ are spawned from the original training-set by coupling the paired training sample $(\mathbf{O},T)$ with pseudo reference $\hat{T}$ taken from the n-best tranpairedions generated by ASR for $\mathbf{O}$. To enable the model to identify mismatched speech and reference (off-task inputs are inevitable in real world), some pseudo references $\hat{T}$ are chosen randomly from the training-set. 
The training data is expanded 30 times in this way. Distribution of the augmented data used in anchor model training is depicted in the lower panel of Fig.\ref{fig:MSE Distribution}. This way of expanding training data  improves data diversity by simulating the errors in transcription from \textit{mispronunciation} and \textit{incorrect} articulation often made by non-native speaker or children speakers. Adopting pseudo-scores makes it possible to train the anchor model in semi-supervised manner using easily obtainable data requiring no extra human effort on data scoring. Data augmentation could be done at much larger scale before deploying model into production. 

\item{\textbf{OOD Testing Data}}

 In real world, 
 subjective uncertainties, human errors, and distribution mismatch are inevitable \cite{knill}.
To investigate the model's robustness in real life, it has to be tested on challenging, unseen samples.

    A five-hour \textit{SingaKids} English corpus \cite{gu20b_interspeech,Chen2016SingaKidsMandarinSC}, including 105 Singaporean children speakers and 2,532 sentences, is graded by tutors 
according to a 5-bands scoring rubric (1 being the lowest and 5 being the highest). Compared to \textit{ SpeechOcean762}, \textit{SingaKids} adopts a different grading scale and targets different age and ethnic group. 
    5 \% of the sentences have mismatched reference due to human mistakes happened in audio recording. This part of the data is identified by graders and its proficiency scores are labeled as the lowest score in marking. Samples from this data-set can correspond to the real-world scenario of encountering unseen data.

\end{itemize}

\subsection{Evaluation Model}
A vanilla implementation of transformer encoder is adopted to map the phone-wise feature sequence to sentence-level proficiency scores. We use the same GOPT configuration as \cite{gopt}. Phone-wise ensemble  features are projected into 24-dimension evaluation embedding. Evaluation embedding sequence plus extra [CLS] tokens are fed into the transformer encoder. Each encoder layer has 8 self-attention heads. And 3 such layers are stacked together. 

Different from original GOPT, encoder outputs at places corresponding to [CLS] are fed into a linear layer followed by a $\mathbf{tanh}$ activation to predict scores. Labels are re-scaled to targets in the range of  $\mathcal{R}_{\mathbf{tanh}}=(-1,1)$ by a transformation.
\begin{equation}
s^{target}=(2\cdot s^{label}-a-b)/(b-a)
\end{equation}
where $[a,b]$ is the range of raw labels. This modification enables the evaluation model to work with different marking scales. According to Eq.\eqref{ni}, pseudo-scores are in the range of $[0,1]$ while human-scores of \textit{SpeechOcean762} are in the range of $[1,10]$. They have to be re-scaled into the same range while training with the iMSE loss in Eq.\eqref{imse}. 
Raw predictions are re-scaled by the inverse transformation at inference time. In cross-data-set test,  raw predictions are re-scaled in the same way to compare with human-scores on OOD data.

Multiple features are encapsulated in the same way as \cite{3m} does. To compare different features and difference training strategies, the following models  and corresponding configurations are investigated in experiment:
\begin{itemize}
\item{\textbf{M-SSL}}\\
Its input is a 3152-dimension ensemble feature, including classical speech evaluation features: $GOP$, $Tempo$, $PhoEmb$, $Pitch$, and 3 self-supervised learning (SSL) speech embeddings, $\mathcal{H}_{wav2vec2}$, $\mathcal{H}_{hubert}$, and $\mathcal{H}_{wavlm}$.
The model has 87k parameters.
It is trained using standard MSE loss on \textit{SpeechOcean762}.
\item{\textbf{M-SSL/WSL}}\\
This model has the same configuration as M-SSL and is trained on the same data and using the the same MSE. The only difference between them is the additional weakly supervised learning (WSL) feature  $\mathcal{H}_{whisper}$ extracted using Whisper model. Its input feature size is  increased to 4432-dimension. It has 117k parameters.
\item{\textbf{M-iMSE}}\\
It uses the same ensemble feature and has the same model structure as M-SSL/WSL. It is trained by the new two-stage method. In the first stage, a separate anchor model with exactly the same structure is trained using pseudo data (introduced in Sec.\ref{ssec:data}) and standard MSE loss.  Pseudo-score labels are computed as the normalized mutual information between ground-truth reference  and pseudo reference (see Sec.\ref{ssec:obj}). In the second stage, a real evaluation model is trained on \textit{speechocean762} using the iMSE loss defined by Eq.\eqref{imse} ($\rho=0.25$ in experiment setting), where the pre-trained anchor model is used to predict the pseudo-scores and to generate the training loss interpolation.
\end{itemize}

\begin{table*}[!ht]
\linespread{1}
\centering
\caption{Performance Comparison by RMSE and PCC}
\vspace*{2mm}
\centering
\begin{tabular}{l|c|c|c|c|c|c|c|c|c|c|c|c}
\hline\hline
\multirow{3}{*}{\textbf{Model}}
 & \multicolumn{6}{c|}{\textit{\textbf{SpeechOcean762}}}
 & \multicolumn{6}{c}{\textit{\textbf{SingaKids}}}\\
 \cline{2-13}
 ~ & \multicolumn{2}{c|}{Pronunciation} 
 & \multicolumn{2}{c|}{Rhythm} 
 & \multicolumn{2}{c|}{Intonation} 
 & \multicolumn{2}{c|}{Pronunciation}
 & \multicolumn{2}{c|}{Rhythm}
 & \multicolumn{2}{c}{Intonation}\\ 
 \cline{2-13}
~ & RMS$\downarrow$ & PCC$\uparrow$ & RMS$\downarrow$ & PCC$\uparrow$ & RMS$\downarrow$ & PCC$\uparrow$ 
& RMS$\downarrow$ & PCC$\uparrow$ & RMS$\downarrow$
& PCC$\uparrow$ & RMS $\downarrow$ & PCC$\uparrow$\\
[0.5ex]
\hline\hline
Anchor Only & 2.840 & 0.709 & 2.987 & 0.670 & 2.737 & 0.684 & 1.930 & 0.481 & 1.911 & 0.456 & 2.055 & \textbf{0.427}\\
\hline
LSTM\cite{Zhang2021MultilingualSE} & 1.158 & 0.667 & 0.979 & 0.729 & 0.985 & 0.726 & 1.202 & 0.428 & 1.316 & 0.443 & 1.499 & 0.236\\
\hline
\textbf{M-SSL} & 1.033 & 0.745 & 0.820 & 0.818 & 0.840 & 0.811 & 1.219 & 0.431 & 1.371 & 0.421 & 1.549 & 0.258\\
\hline
\textbf{M-SSL/WSL} & \textbf{1.010} & 0.759 & \textbf{0.817} & \textbf{0.821} & \textbf{0.835} & \textbf{0.815} & 
1.167 & 0.451 & 1.320 & 0.486 & \textbf{1.442} & 0.312\\
\hline
\textbf{M-iMSE} & 1.032 & \textbf{0.785} & 0.936 & 0.819 & 0.905 & 0.812 & \textbf{1.134} & \textbf{0.516} & \textbf{1.282} & \textbf{0.543} & 1.454 & 0.395\\
\hline

\end{tabular}
\label{tab1}
\end{table*}

\section{Results \& Discussion}

 
Tab.\ref{tab1} compares the above three models with a LSTM model (reported in \cite{Zhang2021MultilingualSE}) on \textit{SpeechOcean762} and \textit{SingaKids}, respectively. Models are evaluated using Root-Mean-Squared-Error (RMSE) and Pearson-Correlation-Coefficient (PCC).

On \textit{SpeechOcean762}, all three models perform significantly better than the previous LSTM model. 
Among them, M-SSL/WSL performs the best. This is in line with expectation as Whisper speech embedding is learned using a very large training data. Overall RMSE for M-iMSE is higher than that for M-SSL and M-SSL/WSL. This performance decrease in RMSE is not a surprise as RMSE is mismatched with the training loss of M-iMSE.
 At the same time, M-iMSE achieves almost the same high PCC level as M-SSL/WSL does. It even makes the top PCC while evaluating pronunciation. To the best of our knowledge, this is the highest PCC ever reported for pronunciation on this task. This proves the effectiveness of the proposed two-stage training on in-distribution data.

On \textit{SingaKids} (out-of-distribution), M-SSL and M-WSL  are not better than LSTM with traditional features. From \textit{SpeechOcean762} to \textit{SingaKids}, PCC of M-SSL is decreased by 42\%, 49\%,and 68\% respectively on 3 proficiency aspects.  It suggests that using the powerful GOPT and advanced embedding features leads to over-fitting and the model doesn't generalize to unseen data. two-stage trained M-iMSE beats M-SSL by 20\%, 29\%,and 53\% PCC increase on \textit{SingaKids}. 

\begin{figure}[ht]
  \centering
  \includegraphics[width=\linewidth, scale=0.2]{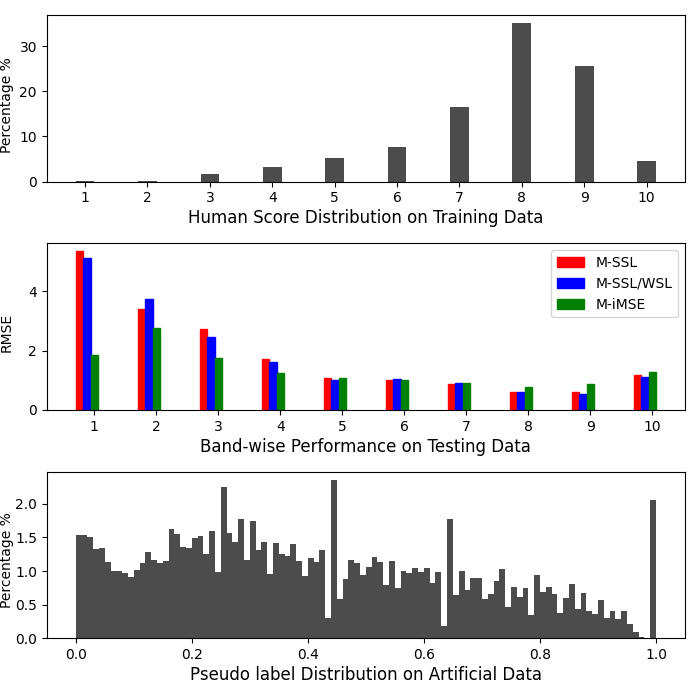}
  \vspace{-0.9 cm}
  \caption{Data vs. RMSE distribution on \textit{SpeechOcean762}.}
  \label{fig:MSE Distribution}
  \vspace{-0.5 cm}
\end{figure}

To gain a deeper insight into the model's performance, band-wise RMSE on \textit{SpeechOcean762} is depicted in the middle panel of Fig.\ref{fig:MSE Distribution}. For M-SSL and M-SSL/WSL, even though both of them achieve very high overall performance on this task, their performance decrease very significantly in under-represented bands (the lower bands in this case). This demonstrates the weakness of standard MSE training on imbalanced data - as the proportion of human labelled training data  (bars in the top) increases, the RMSE of the models (bars in the middle) decreases, and vice versa. Model trained with MSE loss are vulnerable when faced with under-represented data. Observe that band-wise RMSE distribution is the most even for M-iMSE model. M-SSL/WSL performs slightly better than M-SSL. 
However, both of them have wide performance gaps between in-distribution and out-of-distribution tests.

Comparing the results in Tab.\ref{tab1} and Fig.\ref{fig:MSE Distribution}, M-iMSE brings a more balanced band-wise RMSE distribution without sacrificing the overall linear correlation performance (PCC). These results indicate that the proposed two-stage training is an effective approach to improve model robustness when faced with under-represented samples and out-of-distribution samples. It also indicates that mutual information based speech-reference similarity could be a distribution-agnostic learning signal to avoid over-fitting in spoken language assessment.

\section{Conclusion}
Speech evaluation data is often heavily skewed with respect to students' distribution across bands. Models trained on such data face generalization difficulty.
Recently, GOPT and pre-learned speech embedding features brought significant progress for in-distribution speech evaluation. However, the model's robustness has not yet been investigated. In experiment, we found its performance deteriorated seriously when faced with under-represented and out-of-distribution samples. This poses a challenge on consistency and fairness aspects when deploying automatic speech evaluation in real world.

In this work, a mutual-information-based metric is adopted to measure the similarity between speech and reference. An anchor model is pre-trained in semi-supervised manner using pseudo-scores. While training speech evaluation model, prediction made by anchor model is used for loss interpolation to improve robustness. The proposed two-stage training brings more balanced performance across cohorts. Furthermore, new method performs well on out-of-distribution data.

\bibliographystyle{IEEEbib}
\bibliography{strings,refs}

\end{document}